\documentclass[runningheads]{llncs}
\usepackage[T1]{fontenc}
\usepackage{float}
\usepackage{textcomp}
\usepackage{amsmath}
\usepackage{amssymb}
\usepackage{pgfgantt}
\usepackage{graphicx}
\usepackage{amssymb}
\usepackage{caption} 
\usepackage[british]{babel}

\begin{document}
\raggedbottom
\title{Can a Hebbian-like learning rule be avoiding the curse of dimensionality in sparse distributed data?}

\author{
  Maria Os\'orio\\
  \texttt{maria.osorio@tecnico.ulisboa.pt}
  \and
    Luís Sa-Couto\\
  \texttt{luis.sa.couto@tecnico.ulisboa.pt}
  \and
  Andreas Wichert\\
  \texttt{andreas.wichert@tecnico.ulisboa.pt}
}
\institute{Department of Computer Science and Engineering, INESC-ID \& Instituto Superior
Técnico, University of Lisbon, 2744-016 Porto Salvo, Portugal}
\maketitle              % typeset the header of the contribution
\begin{abstract}
It is generally assumed that the brain uses something akin to sparse distributed representations. These representations, however, are high-dimensional and consequently they affect classification performance of traditional Machine Learning models due to ``the curse of dimensionality''. In tasks for which there is a vast amount of labeled data, Deep Networks seem to solve this issue with many layers and a non-Hebbian backpropagation algorithm. The brain, however, seems to be able to solve the problem with few layers. In this work, we hypothesize that this happens by using Hebbian learning.
Actually, the Hebbian-like learning rule of Restricted Boltzmann Machines learns the input patterns asymmetrically. It exclusively learns the  correlation between non-zero values and ignores the zeros, which represent the vast majority of the input dimensionality. By ignoring the zeros “the curse of dimensionality” problem can be avoided.
To test our hypothesis, we generated several sparse datasets and compared the performance of a Restricted Boltzmann Machine classifier with some Backprop-trained networks. The experiments using these codes confirm our initial intuition as the Restricted Boltzmann Machine shows a good generalization performance, while the Neural Networks trained with the backpropagation algorithm overfit the training data.
\end{abstract}
\section{Introduction}
\label{1}

Deep Learning enables high-level abstractions in data using architectures composed of multiple non-linear transformations. These models have significantly improved the state-of-the-art in different domains, such as speech recognition, visual object recognition, object detection and many others \cite{lecun1998handbook,huang2019deep}. However, it is mostly successful in tasks where there are considerable quantities of labeled training data available. Furthermore, the backpropagation learning rule and the high amount of employed layers represents a departure from brain related principles \cite{serre2019deep}. 

The brain is the best natural example of a learning system that can perform extremely difficult tasks in a mostly unsupervised manner \cite{trappenberg2009fundamentals,golomb1990willshaw}. In order to represent information, the brain is thought to share neurons between concepts, which means that a single neuron can be part of the representation of many different concepts. Furthermore, empirical evidence demonstrates that every region of the neocortex represents
information by using sparse activity patterns \cite{ahmad2015properties}.
When looking at any population of neurons in the neocortex their activity will be sparse, that is, a low percentage of neurons are highly active and the remaining neurons are inactive.

Sparse Distributed Representations (SDRs) is the method used to implement computationally the way information is represented in the brain \cite{hinton1984distributed}. An SDR is a binary vector composed of a large number of bits where each bit represents a neuron in the neocortex \cite{ouyang2020learning}. Besides high-dimensional, these vectors are also sparse, which means there is a low percentage of informative (non-zero) bits.

To recognize a particular activity pattern a neuron forms synapses to the active cells in that pattern of activity \cite{palm1980associative}. This way, a neuron only needs to form a small number of synapses, to accurately recognize a sparse pattern in many cells. \cite{BAMI,milnor1985concept,IntroductionTheoryofNeuralComputation}. 

As discussed previously, SDRs are binary vectors composed of many bits, which means we are dealing with a high-dimensional input. These sparse representations are known to work well with associative memory models \cite{palm1982chapter,palm1980associative,golomb1990willshaw,sa2020storing}, but when we try to classify them, we must deal with some problems. 

Classic Machine Learning models, such as, Feed-Forward Networks, are not good at dealing with high-dimensional sparse inputs. The dimensionality of the input data causes the network to have a vast number of parameters, which makes the models prone to overfitting \cite{tan2010learning}. Additionally, the sparseness of the data increases the problem complexity, which affects classification performance due to ``the curse of dimensionality''.

\subsection{Hypothesis}

When dealing with sparse data there is a low percentage of informative bits in the vector. Thus, in order to learn a good and general classifier from these representations, it would be great if there was a model capable of ignoring the empty dimensions and focusing on active units. Ignoring the zeros would ignore a lot of the dimensionality of the sparse vectors, thus solving many of the previously referred issues.

Classic Machine Learning models cannot ignore empty dimensions (zeros) in high-dimensional sparse data. Thus, these models have to learn all the dimensions of the sparse input. 

\begin{figure}[h!]
\centering{}\includegraphics[width=0.9\textwidth]{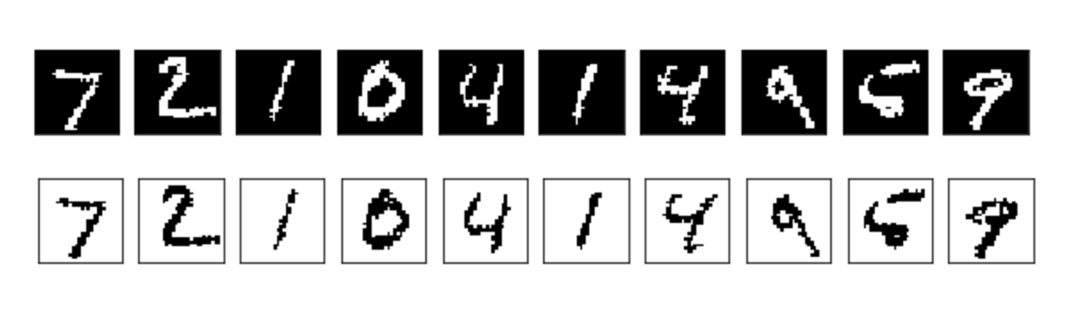}\caption{Ten image sample of MNIST test set. The ten top images represents a binarized version of an original MNIST sample, while the bottom images represent the same sample flipping the bits.}\label{flipped}
\end{figure}

A good way to prove this statement is to classify the same information encoded in two different ways. In Figure \ref{flipped}, the ten top images represent the binarized version of the well-known MNIST\footnote{http://yann.lecun.com/exdb/mnist/} dataset, in which the bits representing each digit are set to 1 and the background information to 0. The ten bottom images represent a flipped sample of the original binarized MNIST, where the bits representing each digit are set to 0 and the background information to 1.

We started by performing the aforementioned experiment using a Logistic Regression (LR), which basically amounts to a softmax-based single layered Neural Network trained by backpropagation of errors and it is generally used as the output layer of a Deep Learning classifier.

By analysing the results on Table \ref{flip-lr}, one concludes that this model has similar performances when classifying the original and the flipped version of the MNIST dataset.

\begin{table}[ht]
\centering
\renewcommand{\arraystretch}{1.2}
\begin{tabular}{c|c|c|}
\cline{2-3}
\multicolumn{1}{l|}{}                                    & \multicolumn{1}{l|}{Train accuracy} & \multicolumn{1}{l|}{Test accuracy} \\ \hline
\multicolumn{1}{|c|}{Original version} & 99.6\%    & 85.2\%                               \\ \hline
\multicolumn{1}{|c|}{Flipped version}       & 99.7\%                               & 84.5\%                              \\ \hline
\end{tabular}
\renewcommand{\arraystretch}{1.2}
\caption{Train and test mean accuracies of LR when classifying a sample of the original binarized MNIST and a flipped version of the same sample.}
\label{flip-lr}
\end{table}

The similar accuracy results achieved by LR in both problems can be explained by the fact that this model learns the information given by 1s in the same way it does with 0s. This implies that, when LR is dealing with high-dimensional sparse inputs, all the dimensions are learnt. 

Our hypothesis states that  a biologically motivated Hebbian-like local learning rule can ignore the empty dimensions (zero values). In fact, the Restricted Boltzmann Machine (RBM) learning rule ignores the zeros present on the input data, which means it exclusively learns the non-zero dimensions. Considering that the empty dimensions represent the vast majority of the input dimensionality, by ignoring the zeros “the curse of dimensionality” problem can be avoided.

We considered a RBM with the same architecture to classify both versions of the binarized MNIST, the original and the flipped one. If our intuition points in the right direction, the model should be able to accurately classify the original version in which the digits information are represented by 1s and fail on the flipped version in which the digits are represented by 0s.

With a sample of 5000 training examples and 1000 test examples of both datasets, Table \ref{flip-rbm} shows the results achieved by the RBM classifier.

\begin{table}[ht]
\centering
\renewcommand{\arraystretch}{1.2}
\begin{tabular}{c|c|c|}
\cline{2-3}
\multicolumn{1}{l|}{}                                    & \multicolumn{1}{l|}{Train accuracy} & \multicolumn{1}{l|}{Test accuracy} \\ \hline
\multicolumn{1}{|c|}{Original version} & 91.6\%                                  & 86.2\%                               \\ \hline
\multicolumn{1}{|c|}{Flipped version}       & 15.5\%                               & 13.8\%                              \\ \hline
\end{tabular}
\renewcommand{\arraystretch}{1.2}
\caption{Train and test mean accuracies of RBM when classifying a sample of the original binarized MNIST and a flipped version of the same sample.}
\label{flip-rbm}
\end{table}

By analysing the results achieved by the RBMs with 500 hidden units, one can conclude that with the original version of the binarized MNIST, the model seems to learn the correlations between active neurons, which represent the digits. As the active bits represent a relatively small percentage of each sample, the model is able to capture the correlations between these active features and have a good generalization performance.

In the flipped version of the binarized MNIST, the RBM fails completely. This may be justified by the fact that this model is unable to catch the correlations between all the active neurons that represent the background. 

This small experiment validates the strong potential of RBMs to deal with high-dimensional sparse inputs. Our hypothesis is that RBMs ignore the zeros and exclusively capture correlations between active units. Could it be that by ignoring the zeros we can avoid ``the curse of dimensionality'' problem when classifying sparse data?  

\section{Background}\label{2}

\subsection{Restricted Boltzmann Machines\label{2.1}}

Restricted Boltzmann Machines (RBMs) were initially invented under the name Harmonium. They are a variant of Boltzmann machines, with the restriction that there is a single layer of $m$ visible units $\mathbf{v}=(v_{1},v_{2},...,v_{m})$ and a single layer of $n$ hidden units $\mathbf{h}=(h_{1},h_{2},...,h_{n})$
with no visible-visible or hidden-hidden connections \cite{salakhutdinov2007restricted}. 

The energy function of a Restricted Boltzmann Machine can be written as 

\begin{equation}
H(v,h)=-\sum_{i=1}^{n}\sum_{j=1}^{m}w_{ij}\cdot h_{i}\cdot v_{j}-\sum_{j=1}^{m}b_{j}\cdot v_{j}-\sum_{i=1}^{n}c_{i}\cdot h_{i}
\end{equation}

For all $i\in{1,...,n}$ and $j\in{1,...,m}$, $w_{ij}$ is a real
valued weight associated with the edge between the units $v_{j}$
and $h_{i}$, and $b_{j}$ and $c_{i}$ are real valued bias terms
associated with unit $j$ of the visible layer and unit $i$ of the
hidden layer, respectively \cite{meyder2008fundamental}.

\subsubsection{Contrastive divergence algorithm}

Obtaining unbiased estimates of the log-likelihood gradient using
Markov Chain Monte Carlo (MCMC) methods typically requires many sampling
steps. However, it has been shown that estimates obtained after running
the chain for just a few steps can be sufficient for model training.

Contrastive divergence (CD) speeds up the computing time of the negative learning phase as it does not use Gibbs sampling to reach thermal equilibrium \cite{fischer2014training}. In this algorithm, the training phase starts by clamping the visible
units with $v^{0}$ and the hidden layer units $h^{0}$ can be computed
by

\begin{equation}
p(h_{i}=1|v)=\sigma\left(\sum_{j=1}^{n}w_{ij}\cdot v_{j}+c_{i}\right)\label{eq:52}
\end{equation}

that define 

\begin{equation}
\left\langle v_{i}h_{j}\right\rangle _{data}^{0}
\end{equation}

As we saw previously there are no visible-visible or hidden-hidden
connections. For that reason each unit $h_{i}$ is independent of
the other hidden units. Therefore, $h^{0}$ can be computed in parallel
as each hidden unit only depends on the visible units connected to
it \cite{hinton2010deep}.

The second step consists in updating all the visible units in parallel
to get a \textquotedblleft reconstruction\textquotedblright{} $v^{1}$,
which can be computed by

\begin{equation}
p(v_{i}=1|h)=\sigma\left(\sum_{j=1}^{n}w_{ij}\cdot h_{j}+b_{i}\right)
\end{equation}

that define 

\begin{equation}
\left\langle v_{i}h_{j}\right\rangle _{recon}^{1}
\end{equation}

The visible units are now clamped with $v^{1}$ and the hidden layer
units $h^{1}$ are computed in parallel using Equation (\ref{eq:52}). 

The reconstruction algorithm can be computed $\tau$ times or until
convergence is reached. Sometimes CD may take many iterations $(1\ll\tau)$
to converge. When $\tau=1$ we are computing a single-step reconstruction
\cite{wichert2020principles}.

The weights update computed for $\tau$ steps of the reconstruction
algorithm is given by

\begin{equation}
\Delta w_{ij}=\eta\cdot(\left\langle v_{i}h_{j}\right\rangle _{data}^{0}-\left\langle v_{i}h_{j}\right\rangle _{recon}^{\tau})
\end{equation}

Hebbian learning is a form of activity-dependent synaptic plasticity where correlated activation of pre and postsynaptic neurons leads to the strengthening of the connection between the two neurons \cite{jaeger2015encyclopedia}. 

The learning rule of RBMs is a type of Hebbian learning as the weights are raised in proportion to the correlations between the states of
nodes \(i\) and \(j\), where the visible vectors are clamped to a vector in the training data and the hidden
states are randomly chosen to be 0 or 1.

\subsubsection{Persistent Contrastive Divergence algorithm}

A different strategy that resolves many of the problems with CD is to initialize the Markov chains at each gradient step with their states from the previous gradient step. This approach was first discovered under the name Stochastic Maximum Likelihood in the applied mathematics and statistics community and later independently rediscovered under the name Persistent Contrastive Divergence (PCD) \cite{younes1998stochastic}.

The idea behind this approach is that, as long as the steps taken by the stochastic
gradient algorithm are small, the model from the previous step will be similar to
the current model. It follows that the samples from the previous
model’s distribution will be very close to being fair samples from the current
model’s distribution.

As each Markov chain is continually updated throughout the learning
process, rather than restarted at each gradient step, the chains are free to wander
far enough to find all the model’s minima. PCD is thus considerably more resistant
to forming models with spurious minima than the original CD algorithm is \cite{mnih2012conditional,goodfellow2016deep}. 

\subsubsection{Restricted Boltzmann Machines for classification\label{proposed architecture}}

In general, RBMs are described and thought of as generative models. However, one can look at the same architecture as feed-forward network classifier with a different learning algorithm.

\begin{figure}[h!]
\centering{}{\includegraphics[width=1\textwidth]{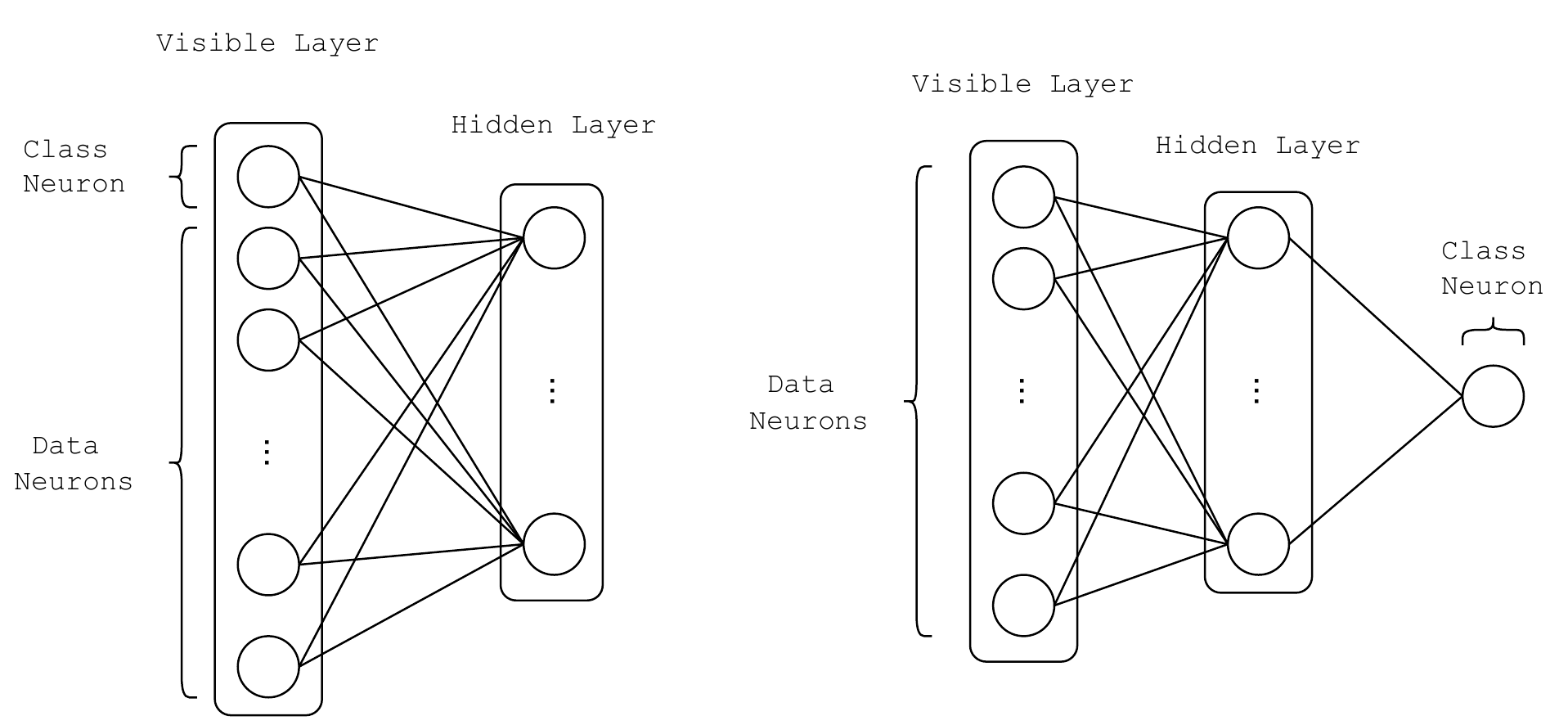}\caption{RBM that models the joint probability distribution of input images and the corresponding labels (left) and the Feed-Forward perspective of the same network (right).}
\label{FFN-RBM}}
\end{figure}

First we have the training phase, where the RBM learns to model the joint probability distribution of input data (explanatory variables) and the corresponding label (output variable), both represented by the visible units of the model as shown in the left network on Figure \ref{FFN-RBM}. The RBM is trained with one of the previously described algorithm: either CD or PCD.

Following the training phase, we have the sampling where the label corresponding to an input example can be obtained by fixing the visible variables that correspond to the data and then sampling the remaining visible variables allocated to the labels from the joined probability distribution of data and labels modeled by the RBM. Hence, a new input example can be clamped to the corresponding visible neurons and the label can be predicted by sampling \cite{fischer2014training}.

\subsection{Backpropagation Neural Networks}

A Deep Learning architecture is a multi-layer stack of simple modules (layers), many of which compute non-linear mappings \cite{lecun2015deep}. The basis of these complex multi-layered networks is a single layered model called Logistic Regression, which is often described as the output of those multi-layered architectures. These networks can be trained by using Stochastic Gradient Descent. As long as the neurons at each layer are relatively smooth functions of their inputs and of their internal weights, one can compute gradients using the backpropagation procedure. This algorithm can be described by running two main phases iteratively. First, the forward phase propagates the input data throughout the network in order to get an output. Second, the gradient of the objective with respect to the input of a layer can be computed by propagating backwards from the gradient with respect to the output of that layer. The backpropagation equation can be applied repeatedly to propagate gradients through all layers, starting from the output at the right (where the network produces its prediction) all the way to the left (where the input data is fed) \cite{lecun2015deep,goodfellow2016deep}. Thus, error computations start in the final layer and flow backwards, leading to the notion of errors backpropagating through the network. 

Once these gradients have been computed, it is straightforward to compute the gradients with respect to the weights between each layer \(k\) with the expression for the partial derivative of the error with respect to each weight:

\begin{equation}
\Delta w_{ij}^{k}=-\eta\cdot\frac{\partial E}{\partial w_{ik}^{k}}\label{backprop}
\end{equation}

Backpropagation learning rule (Eq. \ref{backprop}), adjusts the weights to minimize the error between the actual output and the output predicted by the network. Actually, this is a non-Hebbian learning rule as the weights are not changed considering local correlations between neurons.

\begin{figure}[h!]
\centering{}{\includegraphics[width=1\textwidth]{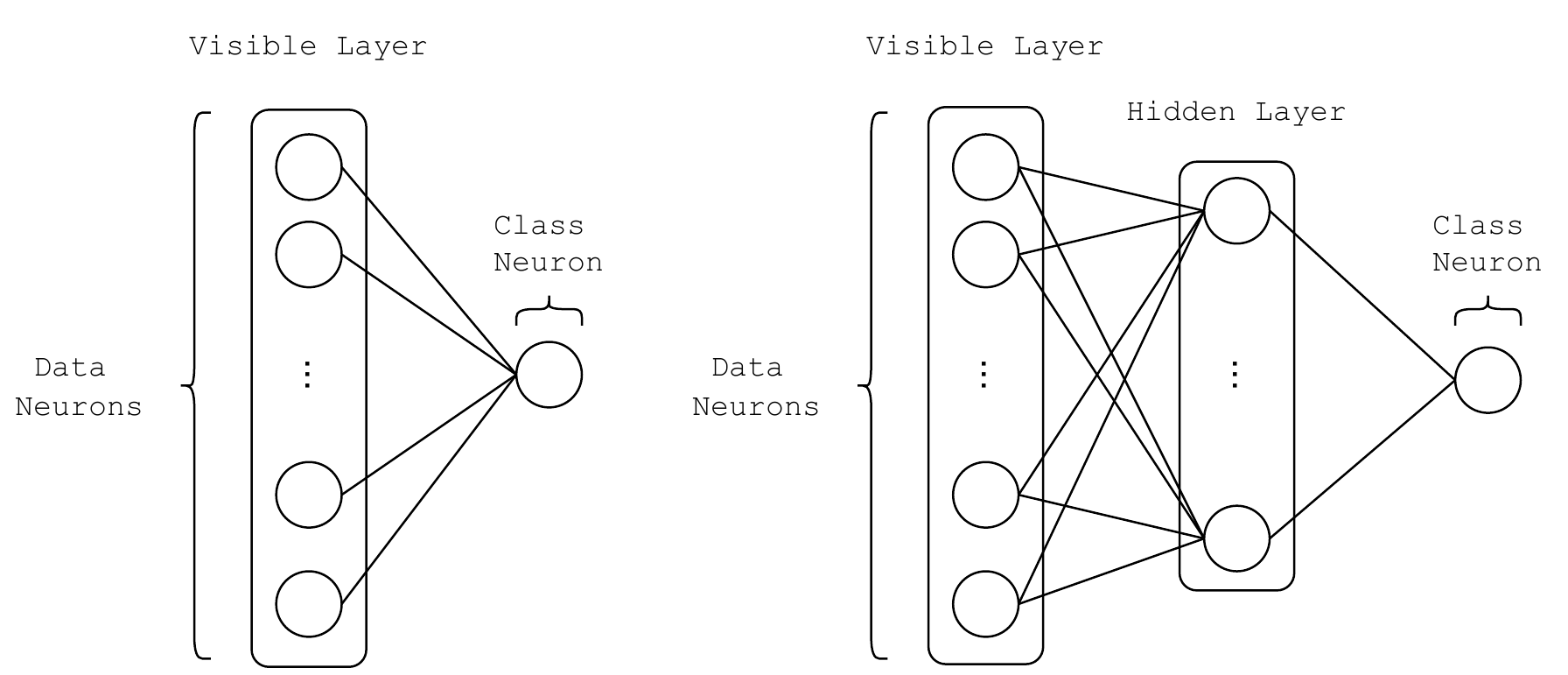}\caption{Architecture of a single-layer Neural Network, which is also known as Logistic Regression (left) and a multi-layer Neural Network with one hidden layer (right).}
\label{lr-rbm}}
\end{figure}

Figure \ref{lr-rbm} shows two architectures of Neural Networks which can be trained with the previously described backpropagation algorithm. The architecture on the left represents a Neural Network without any hidden layers, which can be called a Logistic Regression. The network on the right has the same input and output layers, however it has one hidden layer between them. Many more layers could be added to these Neural Networks, however, the architectures in Figure \ref{lr-rbm} were chosen to be comparable with the Restricted Boltzmann Machine model which has a single hidden layer.

In fact, the only difference between this Neural Network with one hidden layer and the RBM is the learning algorithm. While the Neural Networks in Figure \ref{lr-rbm} are trained with backpropagation of error, where the weights are adjusted to minimize the error between the actual output and the predicted output of the network. The RBM learns using the Contrastive Divergence algorithm which has a local Hebbian-like learning rule, where there is no error propagation, instead it learns to model the joint probability distribution of input data and the corresponding label.

\section{Experiments}\label{3}

With the desire to understand if our intuition pointed in the right direction, several experiments were carried out. 

The first step required to perform these experiments is the definition of the used sparse data. In fact, generating binary sparse data is not a trivial task as it is hard to find a complexity balance in the learning problem. Thus, instead of generating binary data, we decided to generate a dataset where each class follows a multivariate normal distribution.

Bearing in mind our hypothesis, we started by implementing a RBM. This implementation is modelled with Bernoulli visible and hidden units, which means that this RBM is prepared to receive input data in the range [0,1]. By considering that the generated data is real-valued, some exploratory experiments with a Gaussian-Bernoulli RBM were performed, but tuning the value of the Standard Deviation parameter is a hard task, which can produce an unstable learning process \cite{hinton2012practical}. 

In fact, the only difference between using Bernoulli or Gaussian visible units occurs when sampling the visible units of the negative phase of the learning algorithm. Constraining the values of the visible units to be between 0 and 1 imparts a kind of regularization to the learning process. In the sampling phase, the use of Bernoulli visible units is necessary as the sampled label is binary.

\subsection{Dataset generation\label{dataset generation}}

Each dataset is generated with two classes, in which each one follows a Gaussian distribution. The first half of the samples belong to class 0 and follow a Gaussian distribution with mean centered at the origin, the other half corresponds to class 1 and follows another Gaussian distribution with mean centered in a vector of fives. The covariance matrix for each class is defined as a diagonal matrix in which the diagonal values are set to the norm of the difference between the mean vector of each class multiplied by a small number as to reach a good balance in problem complexity.

Moreover, the number of features will be further defined for each experiment. In case the dataset is intended to be dense, the number of features is set to a low value, whereas in the case where the dataset is intended to be high-dimensional, the number of features is fixed to a high value.

\subsection{Pipeline\label{pipeline}}

For each experiment, we started by setting the parameters and then running 10 times the following pipeline:

\begin{enumerate}
\item Populating the dataset by sampling from the two multivariate normal distribution with the previously defined parameters and associate each multivariate normal distribution to a class, either 0 or 1.
\item Centering the data, which consists in subtracting the mean of each feature to every value of that feature.
\item Transforming the dataset into sparse data, which means choosing a few random features to keep in each sample and set the remaining features to zero.
\item Dividing the samples of the dataset into train and test, with the respective percentages of 80\% and 20\%.
\item Training a LR model with the generated train set.
\item Evaluating a LR model by computing and storing the train and test accuracies.
\item Training the RBM with the generated train set using the PCD algorithm. 
\item Evaluating the RBM by performing Gibbs sampling to get the reconstruction of the class unit for both train and test sets. After having all the reconstructions, the model's train and test accuracies can be computed and stored. 
\end{enumerate}

After running the described pipeline, four lists with 10 train and test accuracy values for both models were obtained. Subsequently, the mean and the standard deviation for each list was calculated. In the end, a single train accuracy for both models and a single test accuracy for both models were stored, as well as the respective standard deviations.

\subsection{Experimental Analysis}

We started by generating a dense and a sparse dataset. The dense dataset was generated as described in section \ref{dataset generation}. For the dataset to be dense, the number of features was fixed to 500 and all the values of the data were kept. When generating the high-dimensional sparse dataset, the methodology described in section \ref{dataset generation} was also used. In this case, the number of features was set to 5000. Furthermore, the sparsity was fixed to 95\%, which means that for each sample 5\% of the features were kept and the remaining values set to 0. 

\begin{figure}[h!]
\centering{}{\includegraphics[width=0.7\textwidth]{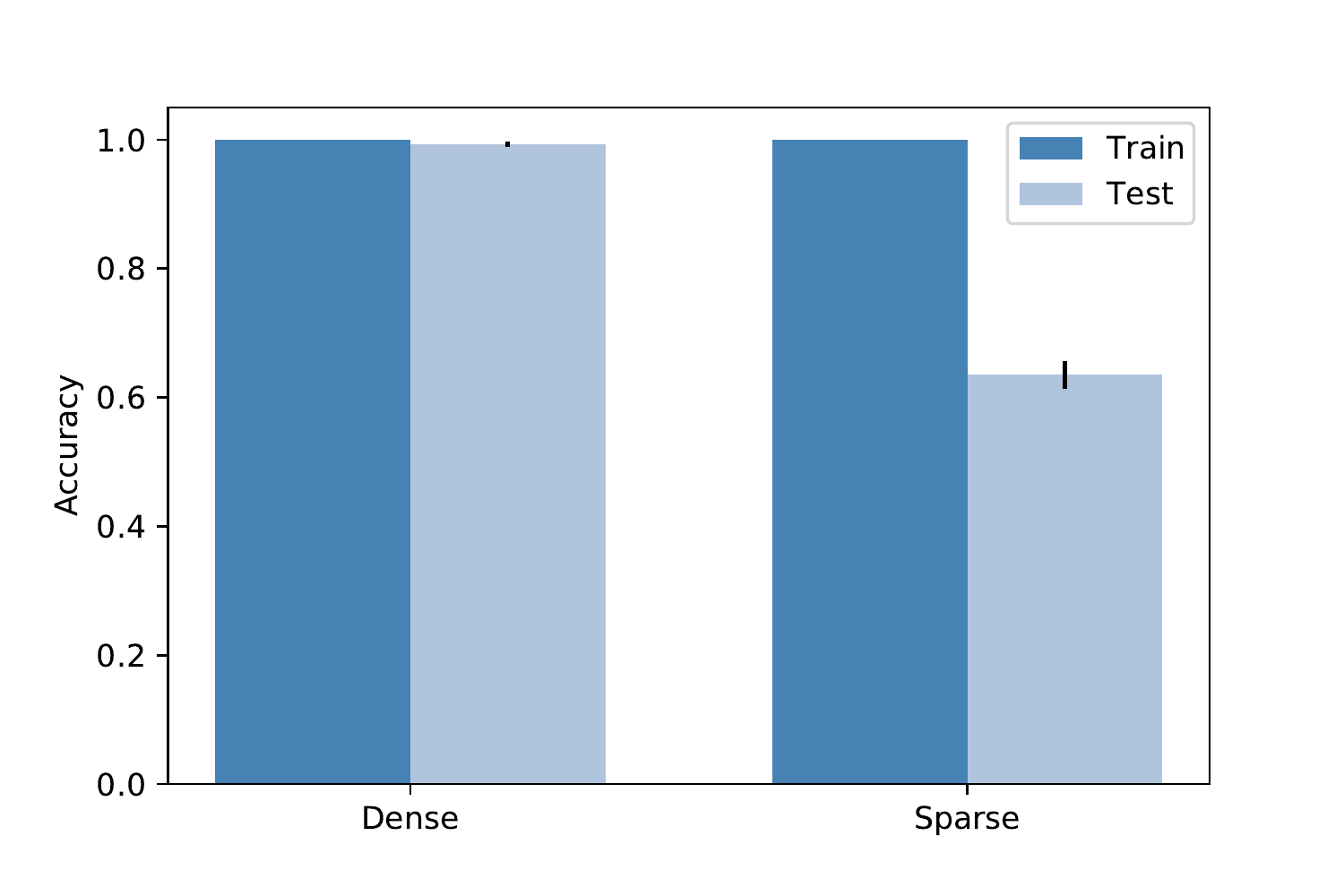}}
\caption{Performance of the LR in a dense versus a high-dimensional sparse dataset.}
\label{dense-sparse}
\end{figure}

In Figure \ref{dense-sparse}, the results show that LR performs well when the dataset is dense, with a mean train accuracy of 100\% and a mean test accuracy of 99.25\%. However, when analysing its performance on the high-dimensional sparse dataset the model registers tremendous overfitting, with a mean train accuracy of 100\% and a mean test accuracy of 63.5\%.

This first experiment provides a baseline to guide the next steps. In what follows, the intent was to show that the RBM performs accurately in a classification task with high-dimensional sparse data. Before diving into the experiments, the parameters must be defined. With these experiments, the aim was to assess the behaviour of a LR and a RBM with increasing sparseness of the dataset. For this reason, the remaining parameters of both models were fixed to the same values, as to achieve a trustful comparison between models. 

The number of samples was fixed to 2000 and the dimensionality of the input to 5000. As far as the parameters of the RBM architecture were concerned, the number of hidden units was set to 500. Additionally, a batch size of 50 and a learning rate of 0.1 was used.

\begin{figure}[h!]
\centering{}{\includegraphics[width=0.7\textwidth]{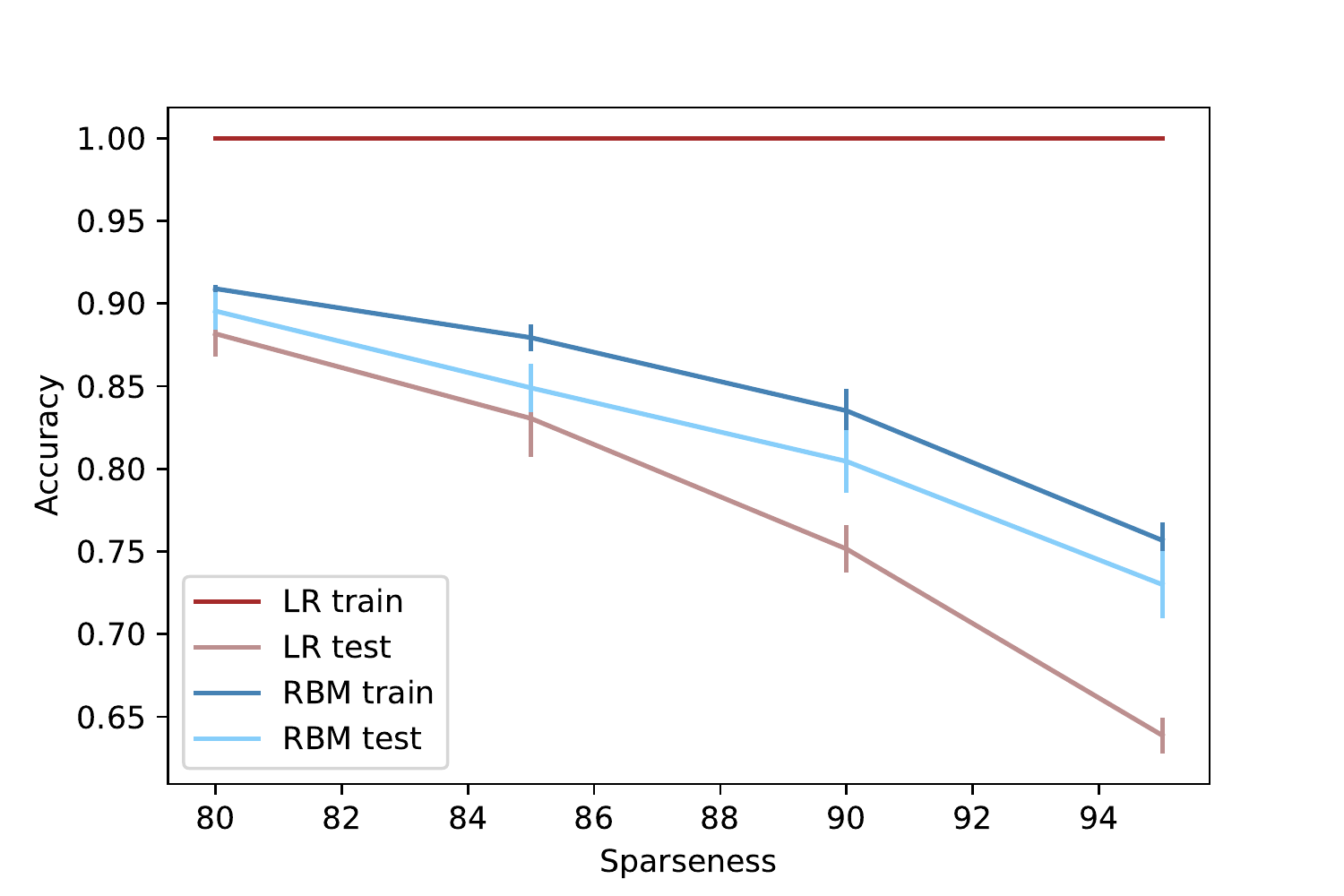}}
\caption{Comparison between accuracies of LR and RBM classifiers with increasing percentage of sparseness.}
\label{rbm-lr-gaus}
\end{figure}

With the final objective of taking meaningful conclusions about both models when the dataset sparsity increases, i.e., the number of zero values increases, the pipeline described in section \ref{pipeline} was followed. To make an easier comparison between models, the sparsity percentage was set to each x-axis value and the mean accuracies of LR and RBM were plotted in Figure \ref{rbm-lr-gaus}.

By analysing the results plotted in Figure \ref{rbm-lr-gaus}, one can observe that the LR classifier has an accuracy of 100\% on the training set, though it is not able to perform accurately on the test set, which suggests that this model is learning the noise in the training data. As the data gets sparser the learning problem becomes harder and test set accuracy decreases. This means that LR can represent the training set of sparse data perfectly, however, is unable to generalize, which results in a poor test set performance. On the contrary, the RBM classifier can generalize the learning problem.

\begin{figure}[h!]
\centering{}{\includegraphics[width=1\textwidth]{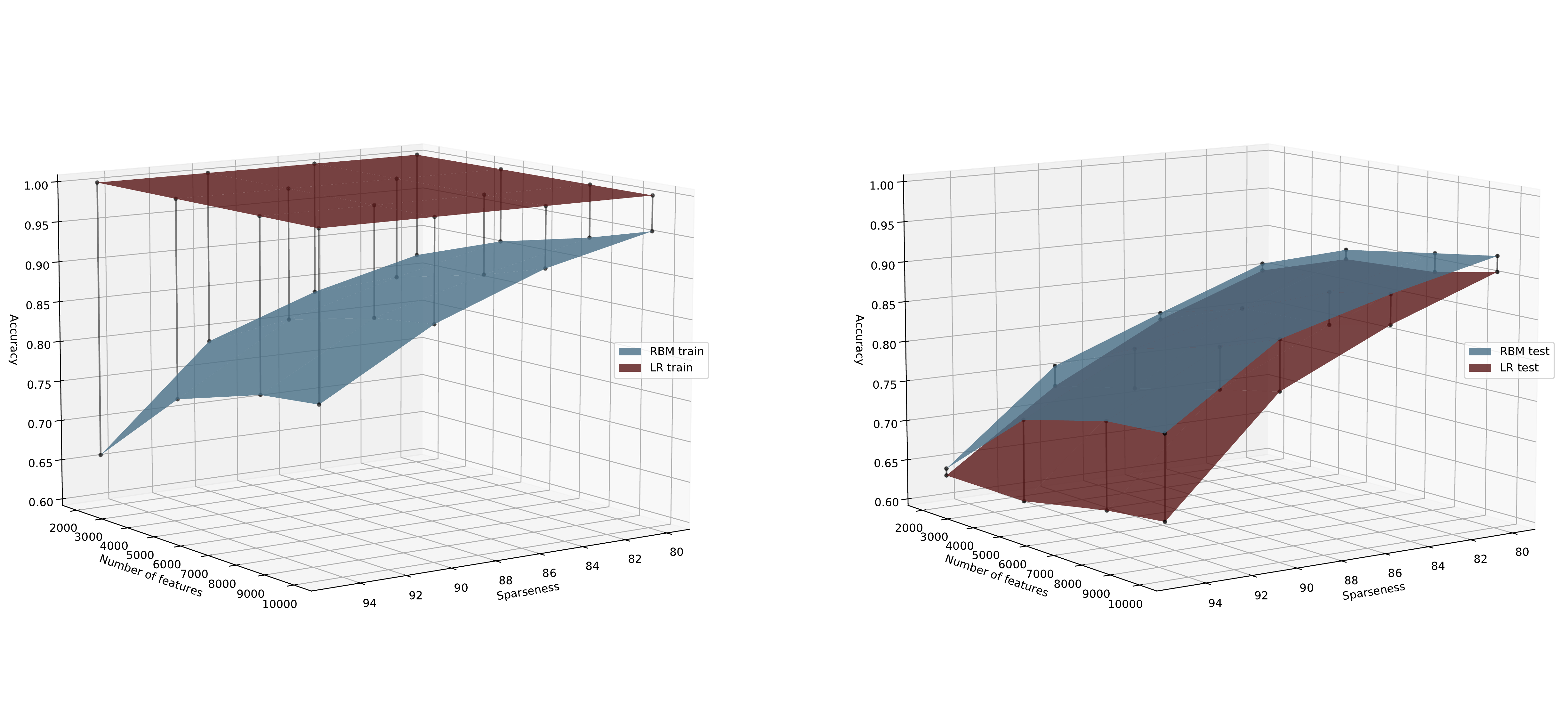}}
\caption{The left plot shows the difference between train accuracies of RBM and LR considering increasing dimensionality and sparseness percentage. The right plot shows the difference between test accuracies of RBM and LR considering increasing dimensionality and sparseness percentage.}
\label{3d}
\end{figure}

To understand how LR and RBM behave with different dimensionalities and sparseness percentages, both models varying these two parameters were run. The 3-D surface plots in Figure \ref{3d}, show the difference between train and test accuracies of RBM and LR considering the number of features (dimensionality) and the sparseness percentage used with each different dimensionality dataset. In the 3-D surface plot on the right, one can observe that as the data gets sparser the difference between both surfaces, which represent the test accuracies, increases and the RBM shows to perform better than the LR. With lower sparseness the RBM and the LR have similar test results. The dimensionality behaves in the same way, which means that with higher dimensional data it is easier to notice the better results achieved by the RBM.  
The 3-D surface plot on the left shows the train accuracy difference of both models. The LR has a train accuracy of 100\%, regardless of the sparsity or the dimensionality of the data. The surface plotted representing the RBM train accuracy is very similar to the RBM test surface, which shows a good generalization performance.

The good results and generalization performance achieved by the RBM classifier can be mainly justified by the fact that this model exclusively learns the correlations between active units and ignores the zeros, which represents most of the dimensionality of the input. Besides, the RBM has a hidden layer that represents hidden correlations between active features of sparse vectors. Therefore, this model can map a high-dimensional sparse vector into a lower dimensional hidden layer, which catches the relevant features present on the high-dimensional sparse vector.

This justification seems to be well grounded, although one may still wonder: Can the good performance of the RBM be justified by the presence of hidden units, which makes it a non-linear classifier? 

To answer this question, a comparison was made between the performances of the RBM and a Multi-Layer Perceptron (MLP). To derive meaningful conclusions, a RBM and a MLP with the same number of hidden units were defined. Additionally, the MLP activation function for the hidden layer units used was the logistic sigmoid function.

More experiments were carried out, in which normally distributed datasets were created following the pipeline described in section \ref{pipeline}. However, instead of comparing the RBM to a LR, the comparison was made with a MLP. In this experiment, the number of samples was fixed to 2000, the dimensionality of the input to 5000 and the number of hidden units of the RBM and the MLP is fixed to 500. With these parameters and a sparsity of 95\%, the mean train and test accuracies in Table \ref{mlp} were obtained.

\begin{table}[h!]
\centering
\renewcommand{\arraystretch}{1.2}
\begin{tabular}{c|c|c|}
\cline{2-3}
\multicolumn{1}{l|}{}                                    & \multicolumn{1}{l|}{Mean train accuracy} & \multicolumn{1}{l|}{Mean test accuracy} \\ \hline
\multicolumn{1}{|c|}{Multi-Layer Perceptron (MLP)} & 100\%                                  & 64.12\%                               \\ \hline
\multicolumn{1}{|c|}{Restricted Boltzmann Machine (RBM)}       & 76.18\%                               & 73.98\%                              \\ \hline
\end{tabular}
\renewcommand{\arraystretch}{1.2}
\caption{Train and test mean accuracies of MLP and RBM given the generated data.}
\label{mlp}
\end{table}

As a matter of fact, the overfitting undergone by the MLP model led to the conclusion that, the generalization capability of the RBM is not a consequence of it being a non-linear classifier. The insertion of a hidden layer does not changes the conclusions taken with the LR. Thus, our initial intuition pointed in the right direction and one can conclude that the main reason for the good generalization performance of the RBM is justified by the Hebbian-like learning algorithm and not by the presence of hidden layers.

\section{Conclusion}
\label{sec:concl}

SDRs are the fundamental form of representing information in the brain. The activity of any population of neurons in the neocortex is sparse, where a low percentage of neurons
are highly active, and the remaining neurons are inactive \cite{BAMI}.
Previous research explored these representations with biologically plausible models to perform associative memory tasks.
To learn a good and general classifier without running into
``the curse of dimensionality'' problem is a hard task. Deep Networks with non-Hebbian learning progressively reduce the dimensionality of the SDR from layer to layer and have success in tasks in which there is a great amount of data with labels.

The present article explores the capabilities of a Hebbian-like learning rule to side step the limitations that classic Machine Leaning models have, when classifying high dimensional sparse data.

Our hypothesis grounds on the capability of the Hebbian learning of RBMs to ignore the empty dimensions of the input data focusing on active units. 

To analyze the validity of our hypothesis, we defined a dataset generation strategy in which each class followed a multivariate normal distribution. Several experiments were carried out using this high-dimensional sparse data, in which datasets with varying dimensionalities and sparseness were generated.

The good generalization capability achieved by the RBM shows that the local Hebbian-like rule used in the network learning algorithm represents the key factor to avoid ``the curse of dimensionality'' problem in sparse distributed data. This model by exclusively learning the correlation between non-zero neurons is able to map a high-dimensional sparse vector into a hidden layer, which catches the relevant features present on the high-dimensional sparse vector. 

On the other hand, both LR and MLP, by using the non-Hebbian backpropagation algorithm, have to learn all the dimensions of the sparse data and consequently become too adapted to the training set, leading to the overfitting problem. 

Considering that we are dealing with biological-like representations, it is expected that a biologically plausible algorithm can better deal with these representations. We can conclude that using a Hebbian-like learning rule represents a clear advantage when dealing with high-dimensional sparse data. 

\bibliographystyle{splncs04}
\bibliography{references}
\end{document}